%% file: main.tex
\documentclass{article}

\usepackage{microtype}
\usepackage{graphicx}
\usepackage{subfigure}

\usepackage{hyperref}

\usepackage{color} 
\usepackage{xcolor}
\def\lkyy{\textcolor{black}}
\def\lky{\textcolor{black}}
\usepackage[accepted]{icml2024}

\input{def}

\usepackage{graphicx}
\usepackage{amsmath}
\usepackage{amssymb}
\usepackage{mathtools}
\usepackage{amsthm}
\usepackage{bbding}
\usepackage{xspace}
\usepackage{multirow}
\usepackage{algorithm}
\usepackage{algorithmic}
\usepackage{amsfonts} 
\usepackage{booktabs}
\DeclareMathAlphabet\mathbfcal{OMS}{cmsy}{b}{n}

\newcommand{\topline}{\toprule [0.1em]}
\newcommand{\midline}{\midrule [0.05em]}
\newcommand{\bottomline}{\bottomrule [0.1em]}

\usepackage[capitalize,noabbrev]{cleveref}
\crefname{section}{Sec.}{Secs.}
\Crefname{section}{Section}{Sections}
\Crefname{table}{Table}{Tables}
\crefname{table}{Tab.}{Tabs.}
\usepackage[T1]{fontenc}

\usepackage[textsize=tiny]{todonotes}

\icmltitlerunning{DCIR: Dynamic Consistency Intrinsic Reward for 
Multi-Agent Reinforcement Learning}

\begin{document}

\twocolumn[
\icmltitle{DCIR: Dynamic Consistency Intrinsic Reward for \\
Multi-Agent Reinforcement Learning}

\icmlsetsymbol{equal}{*}

\begin{icmlauthorlist}
\icmlauthor{Kunyang Lin}{scut}
\icmlauthor{Yufeng Wang}{scut}
\icmlauthor{Peihao Chen}{scut} \\
\icmlauthor{Runhao Zeng}{szu} 
\icmlauthor{Siyuan Zhou}{hkust}
\icmlauthor{Mingkui Tan}{scut}
\icmlauthor{Chuang Gan}{umass}
\end{icmlauthorlist}

\icmlaffiliation{scut}{South China University of Technology}
\icmlaffiliation{szu}{Shenzhen University}
\icmlaffiliation{hkust}{The Hong Kong University of Science and Technology}
\icmlaffiliation{umass}{UMass Amherst}
\icmlcorrespondingauthor{Mingkui Tan}{mingkuitan@scut.edu.cn}

\icmlkeywords{Machine Learning, ICML}

\vskip 0.3in

]

\printAffiliationsAndNotice{} 

\begin{abstract}
Learning optimal behavior policy for each agent in multi-agent systems is an essential yet difficult problem. Despite fruitful progress in multi-agent reinforcement learning, the challenge of addressing the dynamics of whether two agents should exhibit consistent behaviors is still under-explored.
In this paper, we propose a new approach that enables agents to learn whether their behaviors should be consistent with that of other agents by utilizing intrinsic rewards to learn the optimal policy for each agent. We begin by defining behavior consistency as the divergence in output actions between two agents when provided with the same observation. Subsequently, we introduce dynamic consistency intrinsic reward (DCIR) to stimulate agents to be aware of others' behaviors and determine whether to be consistent with them. Lastly, we devise a dynamic scale network (DSN) that provides learnable scale factors for the agent at every time step to dynamically ascertain whether to award consistent behavior and the magnitude of rewards.
We evaluate DCIR in multiple environments including  Multi-agent Particle, Google Research Football and StarCraft II Micromanagement, demonstrating its efficacy.
\end{abstract}

\section{Introduction}
\label{Intro}

Multi-Agent Reinforcement Learning~(MARL) has been evidenced as an important technique in a wide range of practical real-world tasks. These tasks are set in multi-agent systems with the goal of cooperation, such as robotic control~\cite{NEURIPS2021_65b9eea6,kober2013reinforcement,lillicrap2015continuous}, video games~\cite{overcooked,vinyals2019alphastar,brown2019superhuman}, and autonomous vehicles~\cite{autonomous_survey,shalev2016safe}. In MARL, each agent is a reinforcement learning system with its own perception and decision-making capabilities. The agents are required to collectively learn and optimize their behavior strategies by interacting with the environment to achieve a common goal.

\begin{figure}[t]
    \centering
	\includegraphics[width=1\linewidth]{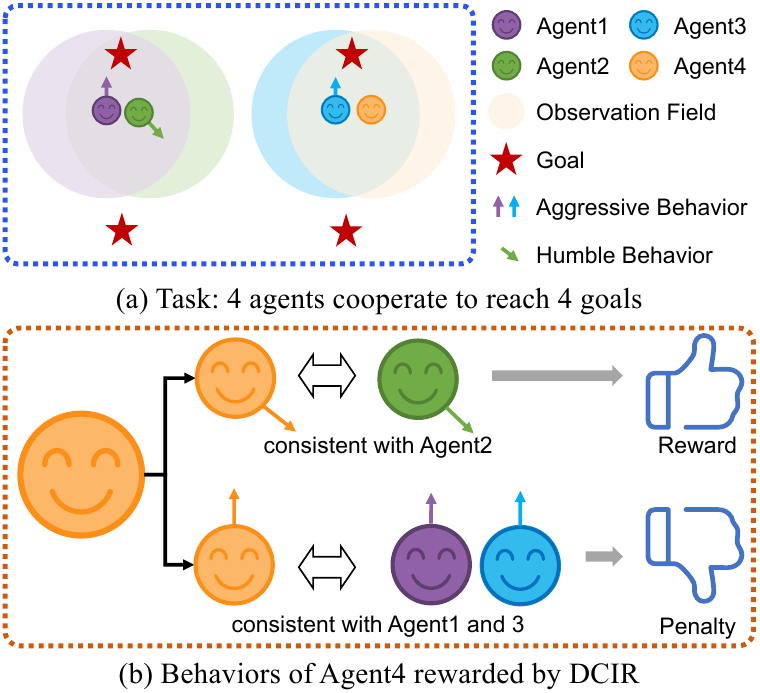}
	\caption{(a) In a multi-agent cooperation task, four agents must cooperate to reach four goals simultaneously as soon as possible. In the context that \textit{Agent1} and \textit{Agent3} take aggressive behavior to directly walk towards the goal they have observed, \textit{Agent2} and \textit{Agent4} should share consistent humble behavior to explore more targets. (b) Our dynamic consistency intrinsic reward (DCIR) awards \textit{Agent4} if it behaves consistently with \textit{Agent2} while punishing it if it behaves consistently with \textit{Agent1} and \textit{Agent3}.
	}
	\vspace{-3mm}
    \label{fig:teaser}
\end{figure}

As the number of agents grows from one to multiple, it is challenging for each agent to find its optimal behavior. This is particularly obvious in many real-world scenarios where extrinsic rewards are sparse and delayed, exacerbating the difficulty of MARL. Deducing the reward of each agent is crucial to encourage the agents to organize their behaviors for successful collaboration~\cite{ELIGN}. 
\lky{While reward shaping~\cite{rs} requires heavy and careful manual work}, previous works are striving to cope with this challenge by assigning intrinsic rewards for each agent~\cite{ELIGN,LIIR,jaderberg2019human,stadie2015incentivizing,iqbal2020coordinated}. LIIR~\cite{LIIR} tries to break down the extrinsic reward into individual learnable intrinsic rewards for each agent.
ELIGN~\cite{ELIGN} encourages multi-agent cooperation through learning a shared world model among different agents.
However, existing methods fail to address the dynamics of whether two agents should exhibit consistent behaviors, which \lky{results in a pitfall in multi-agent reinforcement learning}. 
In this context, consistent behaviors exhibited by two agents denote their tendency to arrive at identical decisions when they are given the same observations. \lkyy{To inspect the existing agents’ ability of dynamic behavior consistency, we conduct a pilot study: to evaluate whether they can behave with 
adequate consistency (see supplementary for more details). We found that existing methods (~\eg ELIGN) put up a poor show: when required to behave consistently, only 75\% agents comply, and even worse in being asked to behave inconsistently (69\%). Thus, incentivizing agents to behave with dynamic consistency remains an unresolved issue.}

To illustrate the necessity of both behavior consistency and inconsistency, we consider a collaborative navigation scenario, where four agents aim to collectively reach four goals.
As shown in Figure~\ref{fig:teaser}, \textit{\textit{Agent1}} and \textit{Agent2} share similar observations, wherein they both observe the same goal. However, to facilitate efficient collaborative task completion and avoid conflicts, it is recommended that they behave inconsistently. Expressly, \textit{Agent1} should adopt a more aggressive approach by navigating directly towards the target, whereas \textit{Agent2} should behave more humbly and continue exploring. In a parallel scenario, \textit{Agent3} exhibits similar aggressive behavior as \textit{Agent1}. Therefore, \textit{Agent4} should maintain behavior consistent with \textit{Agent2} by also adopting a humble exploration behavior towards other goals. Consequently, it is intuitive to reward \textit{Agent4} for adopting the same behavior as \textit{Agent2} and penalize it for imitating the behaviors of \textit{Agent1} or \textit{Agent3}.

Motivated by this observation, in this work, we consider exploring a multi-agent system where each individual agent is incentivized to dynamically and selectively behave consistently with other agents. To formulate behavior consistency, \lky{inspired by agent play style diversity~\cite{charakorn2022generating}}, we expose two agents to the same observation and assess the KL divergence of their predicted action distributions. Intuitively, the action distribution reflects an agent's intention. If two agents have consistent behavior, it suggests that they share similar intentions when facing the same situation. We exploit such consistency in behavior as intrinsic rewards to encourage agents to selectively adhere to the behavior of other agents. However, deciding when to behave consistently or inconsistently with other agents is non-trivial. In order to address this challenge, we introduce a dynamic scaling network (DSN) that calculates learnable scale factors for each agent at every time step to dynamically guide whether to award consistent behavior and the magnitude of intrinsic rewards. The decent combination of behavior consistency and scale factor constitutes our proposed dynamic consistency intrinsic reward (\textbf{DCIR}). \lkyy{DCIR alleviates the dynamic behavior consistency issue in existing methods mentioned above by a large margin~(\ie from 75\% to 88\% in consistent behavior case and from 69\% to 97\% in inconsistent behavior case). }

To sum up, our main contributions are as follows:

$\bullet$ \lky{We study the multi-agent reinforcement learning problem from the perspective of behavior consistency between agents. To this end, we provide a novel viewpoint to assess behavior consistency, formulating it as the divergence in output action distributions between agents given the same observation. }

$\bullet$ We propose Dynamic Consistency Intrinsic Reward~(DCIR) for evaluating the goodness of an agent's behavior. This reward dynamically encourages agents to behave consistently or inconsistently to each other, to maximize extrinsic return and thereby coordinate their behaviors.

$\bullet$ Our extensive experiments demonstrate the superior efficacy of the DCIR method, outperforming five baseline methods on three multi-agent benchmarks consistently, including both cooperative and competitive scenarios, within Multi-agent Particle environment~\cite{lowe2017multi}, Google Research Football environment~\cite{football} and StarCraftII~\cite{starcraft}.

\section{Related Work}
\label{RW}

\subsection{Multi-Agent Reinforcement Learning}

The multi-agent environment, where multiple agents are present for interaction and action, is more typical in practical applications~\cite{cao2012overview,bucsoniu2010multi,zhang2021multi,mcarthur2007multi,berner2019dota, MAPPO, JRPO} compared with the single-agent environment. However, multi-agent reinforcement learning~(MARL) methods are more challenging due to the dependencies and conflicts between multiple agents as well as the dynamic and unstable environment~\cite{alzetta2020time,canese2021multi}. To address this issue, One prominent approach is independent reinforcement learning (IRL)~\cite{zhang2021independent,xu2021stigmergic}, where each agent treats other agents as part of the environment and learns its policy independently. Though simple and intuitive, IRL ignores the dependencies between agents, often leading to sub-optimal outcomes. To capture inter-agent dependencies, centralized training and decentralized execution (CTDE) ~\cite{foerster2016learning,canese2021multi,wang2019influence,lowe2017multi,iqbal2019actor,liu2020pic} is an effective and widely applied framework.  CTDE trains a centralized critic network that observes all agents' joint state and action information during training. During execution, each agent acts independently based on its own observations, leading to decentralized decision-making. CTDE has been widely used in MARL, such as RMIX~\cite{NEURIPS2021_c2626d85}, QMIX~\cite{rashid2020monotonic}, COMA~\cite{foerster2018counterfactual}, VDN~\cite{sunehag2017value}, and QTRAN~\cite{wjkde2019learning}. In this work, we also
follow the CTDE paradigm in our proposed method.

\subsection{Intrinsic Reward in Multi-Agent System}

In a multi-agent system, a single extrinsic reward provided by the environment makes it difficult to coordinate the behavior of multiple agents. Designing intrinsic rewards for each agent is an effective approach to reward allocation in multi-agent environments. In this way, ELIGN~\cite{ELIGN} attempts to encourage agents to match the expectations of their neighbors' expectations and learn collaborative behaviors consistently, making the system more predictable and stable. LIIR~\cite{LIIR} and CDS~\cite{CDS} learn each agent an intrinsic reward function which diversely stimulates the agents at each time step. GIIR~\cite{wu2021generating} uses an intrinsic reward encoder to generate a separate stimulation for each agent.
AIIR-MIX~\cite{li2023aiir} tries to combine intrinsic and extrinsic rewards dynamically in response to changes in the environment. 
Stadie et al.~\cite{stadie2015incentivizing} and Chen et al.~\cite{ActiveCamera} propose a kind of intrinsic reward to encourage each agent to explore the states that are novel to itself, while Iqbal et al.~\cite{iqbal2020coordinated} reward each agent simultaneously to explore states that are also novel to the rest of team agents. However, these intrinsic reward paradigms only promote consistent or diverse behaviors in agents during an episode, but the real world requires agents to autonomously determine when to behave consistently. Our proposed DCIR addresses this issue for better real-world performance.

\section{Preliminaries}
\label{BG}

\subsection{Problem Definition}

    We formulate the MARL problem as a Decentralized Partially Observable Markov Decision Process~(DPOMDP): $<N,\mathbfcal{S},\mathbfcal{O},\mathbfcal{U},\mathbfcal{T},r_\text{ex}>$ \cite{oliehoek2016concise} for $N$ agents in an environment with state space $\mathbfcal{S}$. The observation space for the agents is $\mathbfcal{O} = \left\{\mathbfcal{O}^1,...,\mathbfcal{O}^N\right\}$ and the action space is  $\mathbfcal{U}=\left\{\mathbfcal{U}^1,...,\mathbfcal{U}^N\right\}$. At time step $t$, each agent $i \in N\equiv  \left\{1,...,n \right\} $ obtains its own observation $o^i_t\in \mathbfcal{O}^i$ and performs an action $\boldsymbol{u}^i_t\in \mathbfcal{U}^i$ through the parameterized policy network $\boldsymbol{\pi}_i: \mathbfcal{O}^i\times \mathbfcal{U} \rightarrow \left[0,1\right]$. The environment changes to the next state according to the transition function $\mathbfcal{T}:\mathbfcal{S}\times \mathbfcal{U} \rightarrow \mathbfcal{S}$ with the current state and each agent’s actions and returns a shared extrinsic reward $r_\text{ex}:\mathbfcal{S}\times \mathbfcal{U}\rightarrow \mathbb{R}$. The learning objective of the multi-agent problem is to learn the policy network $\boldsymbol{\pi}_i$ of each agent parameterized by $\theta_i$ to maximize the total expected return: $R=\sum_{t=0}^T \gamma_tr_\text{ex}$, where $\gamma \in \left[0,1\right]$ is the discount factor.

\subsection{Soft Actor-Critic}
We primarily adopt Soft Actor-Critic (SAC) algorithm \cite{sac2018} as the policy optimization method in the experiment. SAC is an off-policy RL algorithm with the Actor-Critic framework. It combines maximizing entropy learning to bring better exploration ability. For each training iteration, the policy parameter $\theta_i$ of agent $i$ is optimized by minimizing the objective $L_{\theta^i}$:

\vspace{-4mm}
\begin{equation}
\begin{aligned}
    L_{\theta^i} = \underset{o_t^i \sim D^i}{\mathbb{E}}\biggl[&\underset{\boldsymbol{u}^i_t \sim \boldsymbol{\pi}_{\theta_i}(\cdot|_t)}{\mathbb{E}}\biggl[\omega \log \boldsymbol{\pi}_{\theta_i}(\boldsymbol{u}^i_t \mid o_t^i) \\
    &-\min_{\phi_i} Q_{\phi_i}^{Soft}(o_t^i, \boldsymbol{u}^i_t)\biggr]\biggr]
\end{aligned}
\end{equation}
\vspace{-4mm}

where $o_t^i$ is the observation of agent $i$ uniformly sampled from the replay buffer $D^i$ at time step $t$; $\boldsymbol{\pi}_{\theta_i}(o_t)$ is the predicted action probability distribution by policy $\boldsymbol{\pi}_{\theta_i}$ input by observation $o_t$; $\omega$ is the entropy temperature coefficient. It determines the importance of entropy maximization in the objective function;  $Q_{\phi_i}^{Soft}(o_t^i, \boldsymbol{u}^i_t)$ is the output action value of the soft value function with parameter $\phi_i$. \lky{In order to alleviate overestimating problem}, two value networks with the same structure are constructed  and the smaller output value of the two networks is chosen during training, that is
\begin{equation}
Q_{\phi_i}^{Soft}(o_t^i, \boldsymbol{u}^i_t) = \min\left\{Q_{\phi_i^1}^{1}(o_t^i, \boldsymbol{u}^i_t), Q_{\phi_i^2}^{2}(o_t^i, \boldsymbol{u}^i_t)\right\}
\end{equation}

\begin{figure}[t]
    \centering
	\includegraphics[width=1\linewidth]{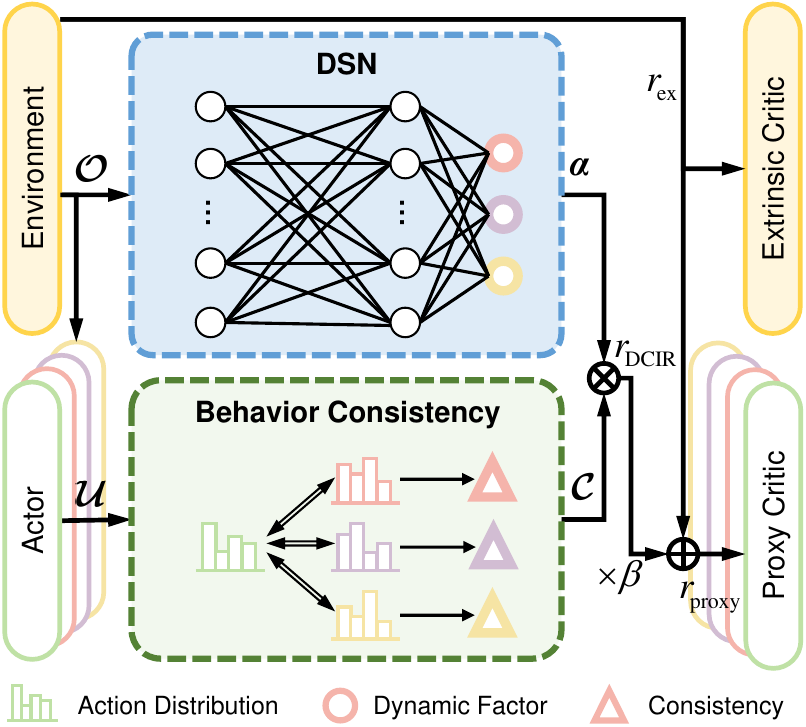}
	\vspace{-5mm}
	\caption{Schematic of the overall DCIR framework. We adopt a bi-level actor-critic framework. Each agent has its own actor and proxy critic (denoted as different colors). The centralized extrinsic critic is used for updating the proposed DSN. Taking the \textcolor[RGB]{191,225,165}{green agent} as an example, at each time step, the Behavior Consistency module calculates its behavior consistency $\mathbfcal{C}$ with the other agents. The DSN outputs the dynamic adjustment factor $\boldsymbol{\alpha}$. Then, the intrinsic reward $r_\text{DCIR}$ is obtained from multiplying the $\mathbfcal{C}$ and $\boldsymbol{\alpha}$. Last, $r_\text{DCIR}$ is added to extrinsic reward $r_\text{ex}$ for proxy value.}
    \label{fig:overview}
    \vspace{-2mm}
\end{figure}

\section{Method}
\label{ME}

In this section, we present a formal description of our proposed \textbf{DCIR} that aims to encourage the agents to adaptively adjust their behavior based on behavior consistency with other agents for finishing multi-agent tasks efficiently. We begin by defining behavior consistency between two agents. Next, we introduce the DCIR framework and its dynamic adjustment approach for each agent. Finally, we formulate an optimizing algorithm for the learning objective.

\subsection{Behavior Consistency between Agents}
\label{BC}

In evaluating the behavior consistency between two agents, our approach involves examining their respective action outputs when confronted with similar states. For analytical purposes, behavior consistency is defined as the degree of similarity in the distribution of actions for two agents given an identical observation.
Specifically, each agent $i$ has its own observation denoted by $o^i_t$ at time step $t$. Since the action space size of each agent is represented as $n$, we denote the action distribution of policy of Agent $i$ as $\boldsymbol{u}^i_t = \boldsymbol{\pi}_i(o^i_t) = \{u_{t, 1}^i,...,u_{t, n}^i\}$. The behavior consistency between agent $i$ and its cooperative agent $j$ is defined by the following KL divergence:

\vspace{-3mm}
\begin{equation}
\mathcal{C}_{i,j,t}= \sum_{k=1}^n u_{t, k}^{i, j} \log \frac{u_{t, k}^{i, j}}{u_{t, k}^i}.
\label{eq:kl}
\end{equation}
\vspace{-3mm}

Here, $u_{t, k}^{i, j}$ represents the probability of agent $j$ selecting action $k$ when it observes the same observation as agent $i$ (i.e., $o^i_t$) at time step $t$. A larger KL divergence (i.e., a larger $\mathcal{C}$) indicates a greater difference in the behaviors taken by the two agents. On the other hand, a smaller KL divergence (i.e., a smaller $\mathcal{C}$) indicates more consistency in the behaviors taken by the two agents. \lky{Note that in continuous action setting, we can still perform KL divergence calculations by replacing the action probability with the mean and variance of the continuous action distribution to sample.}

\subsection{Dynamic Consistency Intrinsic Reward~(DCIR)}
Based on the above analyses, a successful collaboration requires each agent to dynamically 
behave in a consistent or inconsistent manner with other agents. Ideally, the environment should assign a high reward to the agent $i$ when its optimal behavior for the task at time step $t$ is a consistent behavior \wrt agent $j$.
Therefore, an agent's reward at each step is directly related to the difference between its behavior and the behavior of other agents, which is represented by Equation ~\ref{eq:kl}.
To promote the adaptive adjustment of the dominance of consistent and diverse behaviors based on collaboration progress with agent $j$, we propose \textbf{Dynamic Consistency Intrinsic Reward}~(DCIR) for the $i$-th agent as follows:
\begin{equation}
r_{\text {DCIR }}^{i, t} = \sum_{j \in \mathcal{N}(i)} \alpha^i_j \times \mathcal{C}_{i, j, t},
\label{eq:dair}
\end{equation}
 where the value of $\alpha^i_j$ can be negative to encourage consistent behavior between agent $i$ and agent $j$, while positive $\alpha^i_j$ rewards inconsistent behavior. Hence, the introduced adjustment factor $\alpha^i_j$ provides a reliable basis to determine when to reward or punish for the consistent and inconsistent behavior of an agent during collaboration.

\subsection{Dynamic Scale Network~(DSN)}
As aforementioned, factor $\alpha$ is expected to dynamically adjust the consistent reward. To achieve this goal, we propose a \textbf{Dynamic Scale Network} (DSN) to parameterize the learnable $\alpha$. \lky{The DSN is modeled by a multi-layer perceptron (MLP) with layers, each with ReLU activations except the last layer.} For a multi-agent system with $N$ agents, each agent has its own DSN which outputs a vector of length $(N-1)$ for the behavior consistent with other $(N-1)$ agents respectively. Formally, we compute
\begin{equation}
\boldsymbol{\alpha}^i=\operatorname{DSN}\left(o_1, \ldots, o_N\right)
\label{eq:dsn}
\end{equation}
where $\boldsymbol{\alpha}^i =\{\alpha^i_j \mid j \in \mathcal{N}(i)\}$ and $\mathcal{N}(i)$ denotes the set of agents excluding agent $i$. As illustrated in Figure~\ref{fig:overview}, the output $\boldsymbol{\alpha}$ is used to multiply the behavior consistency $\mathcal{C}$ for DCIR in Equation~\ref{eq:dair}.

\begin{algorithm}[!t]
    \small
    \caption{Training paradigm for DCIR}
    \label{alg:training_algorithm}
    \begin{algorithmic}[1]
    \REQUIRE
    The reply buffer $\mD$, the policy of the agent $\boldsymbol{\pi}_{\boldsymbol{\theta}}$, the Dynamic Scale Network $\operatorname{DSN}_{\boldsymbol{\eta}}$.

    \STATE Initialize the $N$ agent parameters $\boldsymbol{\theta}$ and DSN parameters $\boldsymbol{\eta}$ randomly. 
    \STATE Initialize the replay buffer $\mD$
    
    \WHILE{not converge}
        \STATE Collect samples $\{o^i_t,  o^i_{t+1}, r^t_\text{ex}, \boldsymbol{u}^i_t,\{\boldsymbol{u}^{i,j}_t \mid j \in \mathcal{N}(i)\}\}$ from environments and populate them into $\mD$\\
        \FOR{agent $i = 1,..., N$}
            \STATE \lky{Uniformly} sample  $\{o^i_t, o^i_{t+1}, r^t_\text{ex}, \boldsymbol{u}^i_t,\{\boldsymbol{u}^{i,j}_t \mid j \in \mathcal{N}(i)\}\}$ from $\mD$
            \STATE Compute DCIR $r_{\text {DCIR }}^{i, t}$ according to Equation~\ref{eq:kl}, Equation~\ref{eq:dair} and Equation~\ref{eq:dsn} 
            \STATE Calculate proxy reward $r^{i,t}_\text{proxy}$ via Equation~\ref{eq:rp}
            \STATE Update the sample to $\{o^i_t, o^i_{t+1}, r^t_\text{ex}, , r^{i,t}_\text{proxy}, \boldsymbol{u}^i_t,\{\boldsymbol{u}^{i,j}_t \mid j \in \mathcal{N}(i)\}\}$
            \STATE Update $\theta_i$ using Soft Actor-Critic algorithm
            \STATE Update $\eta_i$ using the extrinsic actor-critic optimization method based on the chain derivation rule

        \ENDFOR

    \ENDWHILE
    \end{algorithmic}

\end{algorithm}

\subsection{Learning Objective}
In this section, we present an approach for utilizing the proposed DCIR in multi-agent reinforcement learning.
For each agent $i$ at each time step $t$, we calculate the DCIR using Equation~\ref{eq:dair} and obtain a proxy reward by taking the extrinsic reward~($r^t_\text{ex}$) and $r_{\text {DCIR }}^{i, t}$ into the consideration:
\begin{equation}
r^{i,t}_\text{proxy} = r^t_\text{ex} + \beta \times r_{\text {DCIR }}^{i, t}.
\label{eq:rp}
\end{equation}
where $\beta$ is a scaling hyper-parameter. Here, as outlined problem definition,  the extrinsic reward $r^{i,t}_\text{proxy}$ is sparse and delayed since it is feedbacked by the environment regarding the completion of the task. Then, $r^{i,t}_\text{proxy}$ is used to update the proxy critic via Temporal Difference(TD)~\cite{TD} learning. The updated proxy critic outputs the proxy Q-value used to modify the actor parameters of the agent. 
 
To maximize the standard accumulated discounted team return from the environment, we adopt the \lky{learning paradigm} from LIIR~\cite{LIIR} to train the DSN parameters using an extrinsic-level actor-critic framework. This framework shares the same actors as the proxy actor-critic framework but maintains a centralized extrinsic critic. We utilize the chain rule to connect the impact of the DSN parameter (\ie~$\eta_i$) changes on the objective $J^{\text{ex}}$ of the extrinsic-level actor-critic in the updated actor parameter (\ie~$\theta_i^{\prime}$) as:

\vspace{-3mm}
 \begin{equation}
\nabla_{\eta_i} J^{\text{ex}}=\nabla_{\theta_i^{\prime}} J^\text{ex} \nabla_{\eta_i} \theta_i^{\prime}
\end{equation}
\vspace{-3mm}

More derivation details can be found in the supplementary material. An overview of the optimization process is presented in Algorithm~\ref{alg:training_algorithm}.

\section{Experiment}
\label{EP}

\subsection{Environments}
To evaluate the efficacy of DCIR, we perform experiments on three popular MARL benchmark environments: Multi-agent Particle Environment (MPE)~\cite{lowe2017multi}, Google Research Football ~\cite{football} and StarCraft II Micromanagement~\cite{starcraft}.

\noindent\textbf{Multi-agent Particle Environment (MPE)}
In this environment, multiple particle agents can move, communicate, observe each other, push other agents, and interact with fixed landmarks inhabiting a two-dimensional world. The action space for each agent is discrete,~\ie~stay or change the velocities of four cardinal directions, while the observation space of each agent is continuous. We \lky{adopt} a partially observable setting, wherein each agent possesses an observation radius. These particles can perceive the positions and velocities of other agents within the radius, as well as the positions of landmarks falling within the same range.

\noindent\textbf{Google Research Football(GRF)}
Google Research Football~(GRF) provides a simulated football environment for training multi-agent players to score against the adversary players in a three-dimensional world. The discrete action set consists of 19 actions including one idle action, eight movement actions, four passing/shooting actions, and six other actions. Each football play is controlled by an agent, observing the ball information, team players' information, adversary players' information, controlled player information, and game mode information.

\noindent\textbf{StarCraft II Micromanagement}
StarCraft II Micromanagement is a strategy game where each agent has various attributes such as health points, weapon cooldown, unit type, last action, and distance to observed units. Agents partially observe units within their view range, \ie~a circular radius. The action space includes 4 move directions, a maximum of $k$ attack actions against enemy units, stop, and no operation.

\subsection{Multi-Agent Tasks}
We evaluate the proposed DCIR on three multi-agent tasks based on the MPE, GRF, and StarCraft II Micromanagement. In order to more comprehensively verify the effectiveness of DCIR, we consider both cooperative and competitive tasks.

\noindent\textbf{\textit{Cooperative Navigation. (Coop Nav.)}}
This task is conducted in MPE. In this task, $N$ agents are required to cooperate to reach $N$ goals as fast as possible and are collectively rewarded based on how many goals are occupied. 

\noindent\textbf{\textit{Heterogeneous Navigation. (Hetero Nav)}} There are $N$ goals that $N$ agents cooperate to reach in MPE. The speeds and sizes of $N$ agents are different. Half of the agents are slow and big, while the other half are fast and small.

\noindent\lky{\textbf{\textit{Physical Deception. (Phy Decep.)}} $N$ agents are rewarded if one of them reaches the goal but penalized if one of their $N/2$ adversaries occupies it. The goal is hidden among $N$ landmarks, known only to agents, not adversaries. Agents must learn to deceive adversaries by covering all the landmarks. This task is based on MPE.}

\noindent\lky{\textbf{\textit{Predator-prey. (Pred-prey.)}} In a randomly generated obstacle-filled environment, $N$ slow adversaries pursue and capture $N$ fast cooperating agents. When an adversary catches an agent, the agent is penalized, and the adversary is rewarded. This task is based on MPE.}

\noindent\lky{\textbf{\textit{Keep-away.}} In this MPE task, there are $N$ agents and $N$ landmarks. One of the landmarks is the goal, known to the agents. $M$ adversaries are present, and the agents are rewarded for pushing them away from the goal. Adversaries can only deduce the goal based on the agents' behavior.}

\noindent\textbf{\textit{Academy 3vs1 with Keeper. (3v1 w/ keeper.)}}
This task is undertaken within the GRF environment. Specifically, three agents are assigned to control three individual players. Their objective is to successfully navigate past a defender and a goalkeeper in order to propel the ball into the goal and score.

\noindent\textbf{\textit{3 Marines vs. 3 Marines.}}
We explore symmetric battle scenarios in StarCraft II involving 3 Marines on each side,~\ie~ \textit{3 Marines vs. 3 Marines}~(3M). In each time step, the agents receive a joint team reward based on their inflicted damage and the damage received from the enemy.

\subsection{Experimental Setups}

    We implement the experiments in MPE by tianshou framework~\cite{tianshou}, GRF environment by rllib framework ~\cite{rllib} and StarCraft II Micromanagement by SMAC framework~\cite{starcraft}. \lky{For a fair comparison, we employ the same experiment setting in MPE and GRF as ELIGN and in StarCraft II Micromanagement as LIIR.} Specifically, for MPE, we train the agents for $100k$ time steps until convergence, while $5M$ time steps for GRF and $3M$ time steps for StarCraft II Micromanagement. Each experiment runs on one NVIDIA A800 GPU.

\subsection{Baselines}
\lky{Our overall main focus is on evaluating the effectiveness of DCIR in intrinsic reward systems. Toward this goal, we choose five popular and solid baselines under the same setting as our method for a fair and comprehensive evaluation.} 

\noindent\textbf{SPARSE~\cite{lowe2017multi}}
This SPARSE baseline learns the policy using only sparse extrinsic rewards returned by the environment.

\noindent\textbf{EXP-self~\cite{stadie2015incentivizing}}
This baseline encourages the agent to explore states that it has not explored before. The intrinsic reward is high for exploring more novel states. It is a classic exploration-based intrinsic reward method in the single-agent domain.

\noindent\textbf{EXP-team~\cite{iqbal2020coordinated}}
In addition to only rewarding its exploration of states that are novel to itself, EXP-team also rewards each agent for simultaneously exploring states that are also novel to the rest of the team agents. The rewards are also provided in the form of intrinsic rewards.

\noindent\textbf{ELIGN~\cite{ELIGN}}
ELIGN proposes to use intrinsic rewards to reward each agent for making decisions that conform to the predictions of other agents, ensuring that the behavior of each agent is predictable.

\noindent\textbf{LIIR~\cite{LIIR}}
LIIR proposes to motivate agents to perform diverse behaviors by using learnable intrinsic rewards without specific practical meanings.

\begin{table*}[!t]
\resizebox{1.0\linewidth}{!}{\begin{tabular}{lcccccccccccc}
\topline
\multirow{2}{*}{Methods}                  &  & \multicolumn{3}{c}{MPE~(Cooperative)}     &  & \multicolumn{5}{c}{MPE~(Competitive)}     &  & GRF~(Competitive)                                                   \\           \cmidrule{3-5} \cmidrule{7-11} \cmidrule{13-13} 
                                          &  & Coop Nav. (5v0)          &  & Hetero Nav. (6v0)      &  & Phy Decep. (4v2)      &  &  Pred-prey. (4v4)        &  &  Keep-away. (4v4)   &  & 3v1 w/ keeper. (3v2)              \\ \midline
SPARSE               &  &   $459.92\pm22.44$       &  &  $616.62\pm25.30$      &  &  $166.89\pm27.72$     &  &  $-28.75\pm7.3$        &  &  $0.752\pm1.82$ &  &  $0.020\pm{0.001}$      \\
EXP-self   &  &   $458.45\pm19.79$       &  &  $702.73\pm18.57$       &  &  $146.55\pm29.05$     &  &  $-25.35\pm6.16$      &  &  $10.52\pm5.48$ &  &  $0.024\pm{0.004}$     \\
EXP-team      &  &   $497.15\pm11.47$       &  &  $695.38\pm12.22$       &  &  $84.66\pm16.94$     &  &  $-17.21\pm8.23$       &  &  $1.40\pm2.06$ &  &  $0.021\pm{0.002}$ \\
ELIGN                        &  &   $498.24\pm9.77$        &  &  $646.70\pm23.25$       &  &  $186.83\pm21.92$     &  &  $-9.14\pm5.57$       &  &  $11.29\pm9.02$ &  &  $0.025\pm{0.001}$   \\
LIIR                          &  &   $495.50\pm8.40$       &  &  $660.92\pm15.71$              &  &  $179.27\pm21.74$     &  &  $-49.79\pm11.57$     &  &  $3.45\pm13.53$ &  &  $0.027\pm{0.003}$   \\
\midline
\textbf{DCIR(Ours)}                       &  & $\textbf{525.92}\pm\textbf{8.99}$  &  &$\textbf{707.52}\pm\textbf{16.70}$    &  &  $\textbf{224.49}\pm\textbf{15.39}$     &  &  $-\textbf{7.91}\pm\textbf{1.88}$        &  &  $\textbf{21.45}\pm\textbf{9.75}$    &  &  $\textbf{0.031}\pm\textbf{0.002}$\\ \bottomline
\end{tabular}
}
\caption{Mean test episode extrinsic rewards and standard errors across different methods. We train all algorithms with 5 random seeds following ELIGN. The numbers in parentheses represent the number of agents and adversaries,~\ie~(Agt \# vs. Adv \#).}
\label{tab:map}

\end{table*}

\subsection{Comparison Results}

\noindent\textbf{Results on Cooperative Tasks}
In Table~\ref{tab:map}, the policy learned with our DCIR beats all the baselines on both \textit{Coop Nav.} task and \textit{Hetero Nav.} task. On \textit{Coop Nav.} task, compared to the exploration-based methods EXP-self and EXP-team, the improvement can be attributed to that instead of blindly pursuing novel states, we encourage agents to selectively adopt exploration behaviors. DCIR also beat ELIGN. \lky{ELIGN} encourages the behaviors of agents to be predictable and aligned with each other. When the behaviors of all agents become easy to predict, they also tend to be the same. This limits the agents to dynamically adjust intentions according to the current state. \lky{Different from ELIGN, }our method \lky{does not simply request all the agents acting consistently but considers both behaviors dynamically conditioned on the states. This is more obvious in the \textit{Hetero Nav.}, where the agents need more moments of inconsistent behavior because of differences in their properties.} Compared to LIIR, DCIR better promotes collaboration and coordination among multiple agents instead of focusing on agents' own learning of an intrinsic reward. Moreover, it is difficult to fully learn an intrinsic reward that has no physical meaning in a complex multi-agent environment from scratch. 

\begin{figure}[t]
    \centering
	\includegraphics[width=1.0\linewidth]{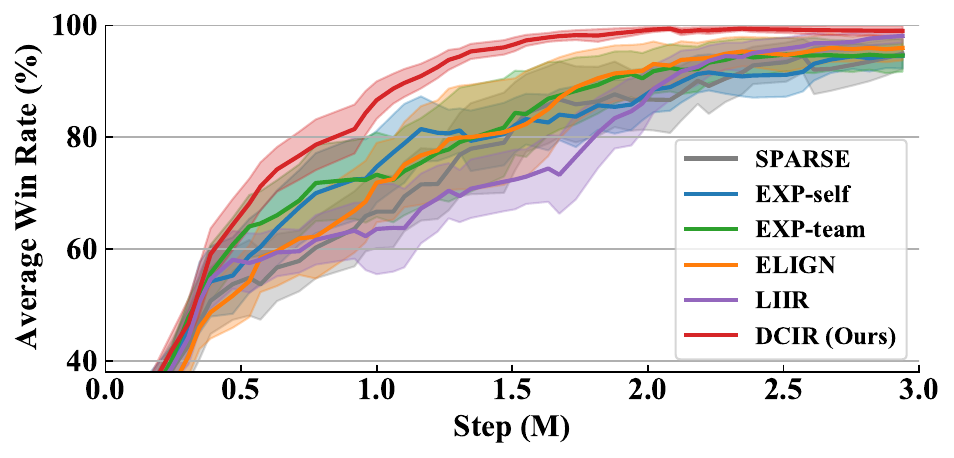}
	\vspace{-5mm}
	\caption{Average test winning rates \vs training steps of various methods on 3M task in StarCraft II. Our training and testing settings remain the same as in the LIIR~\cite{LIIR}, and 5 seeds are selected for plotting.}
    \label{fig:smac}
\end{figure}

\noindent\textbf{Results on Competitive Tasks}
We further demonstrate the effectiveness of our proposed DCIR for tasks with adversaries. The results are shown in Table~\ref{tab:map} and Figure~\ref{fig:smac}. We noticed that DCIR has the highest improvement~(89.99\%) over the SoTA baseline in the \textit{Keep-away.} task. We speculate that the role division is more obvious in this task, \eg~deceiving and pushing, which has high requirements for the behavior consistency between agents and DCIR can alleviate this. Also in the \textit{3v1 w/ keeper.} task, while the baselines have comparable performance, our DCIR can boost the performance by 24\%. This can be explained by the complexity arising from intricate rules, interferences among defenders and keepers, beyond the scope of mere exploration (EXP-self, EXP-team) or alignment (ELIGN) strategies. Additionally, the vast action and state possibilities hinder direct learning of a feasible intrinsic reward (LIIR). Test winning rates in Figure~\ref{fig:smac} on StarCraft II Micromanagement further shows consistent improvement with faster convergence and higher win rate, demonstrating the efficacy of DCIR in strategy competitive MARL task.

\begin{table}[!t]
\centering
\resizebox{1.0\linewidth}{!}{\begin{tabular}{lcccc}
\topline
\multirow{2}{*}{Distance Type} &  &    Cooperative   &  &    Competitive               \\ \cmidrule{3-3} \cmidrule{5-5}
                             &  & Coop Nav. (5v0)  &  & Phy Decep. (4v2)                   \\ \midline
Binary Divergence          &  &  $504.60\pm{7.56}$   &  &  $205.08\pm{14.46}$\\
JS Divergence              &  &  $516.76\pm{3.98}$   &  &  $220.57\pm{13.17}$\\
TV Distance                &  &  $508.27\pm{7.52}$   &  &  $214.64\pm{19.43}$\\ \midline
KL Divergence~(Ours)          &  &  $\textbf{525.92}\pm\textbf{8.99}$ &  &  $\textbf{224.49}\pm\textbf{15.39}$\\ \bottomline
\end{tabular}
}
\caption{Ablation Study on alternatives of KL Divergence.}
\label{tab:disdis}
\end{table}

\begin{table}[!t]
\centering
\resizebox{1.0\linewidth}{!}{\begin{tabular}{lcccc}
\topline
\multirow{2}{*}{Factor Type} &  &    Cooperative   &  &    Competitive               \\ \cmidrule{3-3} \cmidrule{5-5}
                             &  & Coop Nav. (5v0)  &  & Phy Decep. (4v2)                   \\ \midline
Inconsistency                &  &  $491.51\pm{12.16}$   &  &  $197.46\pm{31.92}$\\
Consistency                  &  &  $506.45\pm{8.12}$    &  &  $206.04\pm{19.93}$\\
Shared Factor                &  &  $511.18\pm{25.13}$   &  &  $205.90\pm{21.81}$\\  
Learnable Paras.             &  &  $505.29\pm{12.63}$   &  &  $208.85\pm{17.20}$\\  
\midline
DSN (Ours)                         &  &  $\textbf{525.92}\pm\textbf{8.99}$  &  &  $\textbf{224.49}\pm\textbf{15.39}$\\ \bottomline
\end{tabular}
}
\caption{Ablation Study on the effectiveness of DSN.}
\label{tab:ablation}
\end{table}

\subsection{Ablation Study}
\lky{In this section, we proceed to ablate DCIR under MPE on both cooperative task and competitive task, \ie~\textit{Cooperative Navigation} and \textit{Physical Deception}, respectively.}

\noindent\textbf{Divergence Alternatives}
To quantify the behavior consistency between agents, we calculate the KL divergence between the action probability distributions of two agents when given the same observation. In this section, we \lky{consider three alternatives.} 
For the first variant~(\texttt{Binary Divergence}), the behavior consistency is $+1$ if the agents output the same action otherwise $-1$, indicating their behavior is consistent and inconsistent, respectively. The second variant is a modification of KL divergence, \ie~Jensen-Shannon~(JS) Divergence~\cite{JSD}. The last variant is Total Variant~(TD) Distance~\cite{TVD}, which can be formulated as L1-norm between two distributions.
\lky{In Table~\ref{tab:disdis}, the results suggest that both KL divergence and JS divergence are effective measures for consistency, but KL divergence performs slightly better. The binary consistency method is the worst. We suspect that the binary consistency method is too simplistic and fails to capture the nuances of the action distributions. On the other hand, TV distance is not as effective as KL divergence, which may be due to its sensitivity to small differences between distributions. KL divergence and JS divergence are both smooth measures of the similarity between two probability distributions, which makes them more effective in measuring consistency. }

\begin{figure*}[!t]
    \centering
	\includegraphics[width=1.0\linewidth]{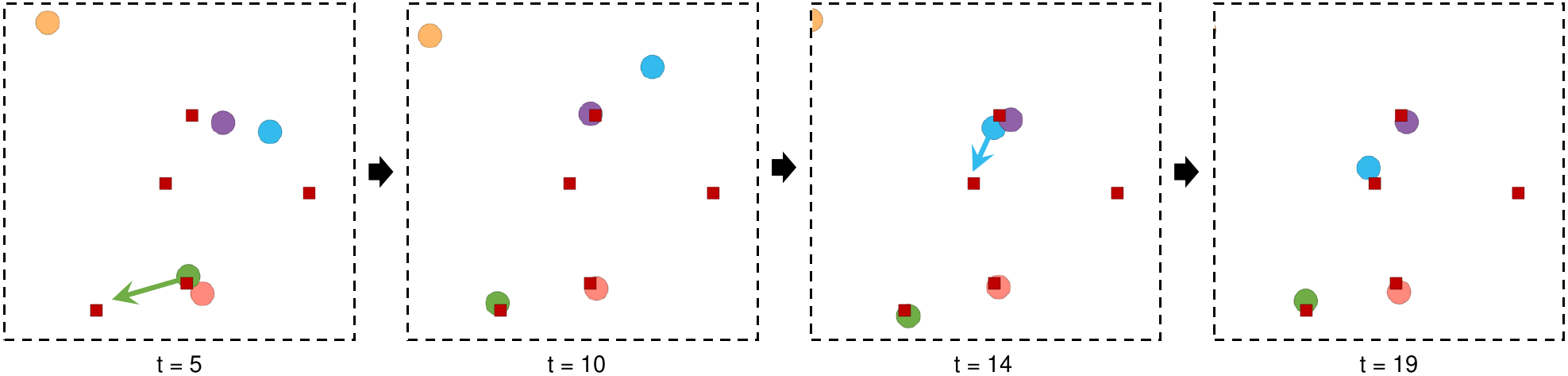}
	\vspace{-5mm}
	\caption{Visualization of \textit{Cooperation Navigation~(5v0)} task. Goal:~\textcolor[RGB]{200,0,0}{$\blacksquare$}. 
    \textit{Agent1}:\textcolor[RGB]{112,173,71}{$\bullet$}.
    \textit{Agent2}:\textcolor[RGB]{255,137,124}{$\bullet$}.
    \textit{Agent3}:\textcolor[RGB]{51,187,238}{$\bullet$}.
    \textit{Agent4}:\textcolor[RGB]{144,97,167}{$\bullet$}.
    \textit{Agent5}:\textcolor[RGB]{255,183,105}{$\bullet$}. 
    Behavior direction of \textit{Agent1}: \textcolor[RGB]{112,173,71}{$\rightarrow$}.
    Behavior direction of \textit{Agent3}: \textcolor[RGB]{51,187,238}{$\rightarrow$.}
    }
	\vspace{-1mm}
    \label{fig:vis_example}
\end{figure*}

\begin{figure}[!t]
	\centering
	\begin{minipage}[t]{.48\linewidth}

        \includegraphics[width=\linewidth]{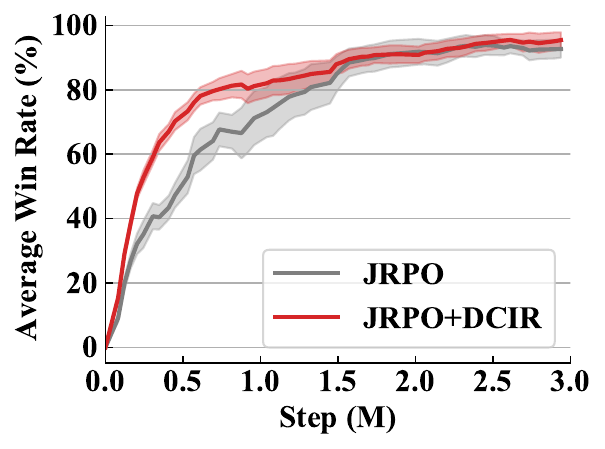}
    \end{minipage}
    \hskip 0.08in
	\begin{minipage}[t]{.48\linewidth}
    	\includegraphics[width=\linewidth]{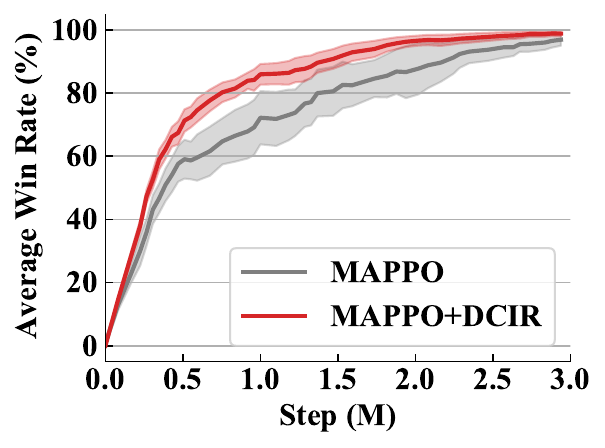}

    \end{minipage}
    \vspace{-5mm}
    \caption{Average test winning rates \vs training steps of incorporating DCIR into MAPPO and JRPO on 3M task in StarCraft II. Our training and testing settings remain the same as in the LIIR~\cite{LIIR}, and 5 seeds are selected for plotting.}
    \label{fig:sota}
\end{figure}


\noindent\textbf{Effectiveness of DSN}
To investigate the effectiveness of the proposed DSN, we replace the output of the dynamically adjustable factor $\alpha$ output from DSN with three variants. We first replace $\alpha$ with $+1$, constructing a variant that encourages an agent to behave inconsistently with any other agent. We thus name it \texttt{Inconsistency}. On the contrary, we replace $\alpha$ with $-1$ to encourage consistent behaviors between agents, and we denote this variant as \texttt{Consistency}. Then, to verify that each agent needs to maintain different behavior consistency with different agents, we design a variant (\ie~\texttt{Shared Factor}). The DSN of each agent in this variant only outputs a shared $\alpha$ for the team agents. Last, we replace $\alpha$ with no-input learnable parameters to justify the necessity of DSN design (\ie~\texttt{Learnable Paras.}).
 As illustrated in Table~\ref{tab:ablation}, either awarding the agents to perform only inconsistent behaviors or consistent behaviors drops the performance. By adding our DSN to learn the dynamic adjustable factor $\alpha$, the agent can learn to tackle different requirements for behavior consistency under each task. Compared to \texttt{Shared Factor}, learning separate $\alpha$ significantly improves the performances. This is because the team agents vary in their policy learning levels and behavior intentions. Therefore, different $\alpha$ needs to be used to encourage different behavior consistency with other team agents. Moreover, the lack of necessary input information and network complexity makes the learnable parameters difficult to distinguish task situations and adjust appropriately. \texttt{Learnable Paras.} thereby performs worse than DSN.
 
\begin{table}[!t]
\centering
\resizebox{1.0\linewidth}{!}{\begin{tabular}{lccccccc}
\topline
\multirow{2}{*}{Method}                      &  &    \multicolumn{2}{c}{Cooperative}                       &  &    \multicolumn{2}{c}{Competitive}                                       \\ \cmidrule{3-4} \cmidrule{6-7}
                                             &  & Coop Nav. (5v0)  &  Coop Nav. (10v0) &  & Phy Decep. (4v2)  &  Phy Decep. (8v4)                 \\ \midline
SPARSE        &  &  $459.92\pm22.44$ & $1103.87\pm{22.89}$   &  &  $166.89\pm27.72$ & $563.42\pm{70.52}$\\
EXP-self      &  &  $458.45\pm19.79$ & $1101.15\pm{42.53}$   &  &  $146.55\pm29.05$ & $334.71\pm{97.83}$\\ 
EXP-team      &  &  $497.15\pm11.47$ & $1106.48\pm{33.14}$   &  &  $84.66\pm16.94$  & $351.02\pm{84.01}$\\ 
ELIGN         &  &  $498.24\pm9.77$  & $1088.42\pm{39.4}$   &  &  $186.83\pm21.92$ & $534.91\pm{48.05}$\\  
LIIR          &  &  $495.50\pm8.40$  & $1003.45\pm{44.1}$   &  &  $179.27\pm21.74$ & $586.83\pm{141.33}$\\ \midline
DCIR          &  &  $\textbf{525.92}\pm\textbf{8.99}$& $\textbf{1137.48}\pm\textbf{47.21}$ &  &  $\textbf{224.49}\pm\textbf{15.39}$ & $\textbf{622.88}\pm\textbf{82.96}$\\ \bottomline

\end{tabular}
}
\caption{Performance when the number of agents increases.}
\label{tab:num}
\vspace{-8mm}
\end{table}

\noindent\textbf{Scalability of DCIR}
In this experiment, we investigate the scalability of DCIR when more agents are added to cooperative and competitive tasks. As shown in Table~\ref{tab:num}, when the number of agents increases, DCIR enjoys superior performance over the five baselines both in the cooperative task and the competitive task. This highlights that DCIR is feasible and effective in the case of a larger number of agents. Furthermore, in contrast to MPE and GRF benchmarks that use SAC, we utilize AC~\cite{ac} algorithm in StarCraft II Micromanagement following LIIR. ~\lkyy{In Fig.~\ref{fig:sota}, we also incorporate our DCIR into advanced MARL algorithms, \ie~MAPPO~\cite{MAPPO} and JRPO~\cite{JRPO}, respectively in StarCraft II Micromanagement. Consistent improvement showcases the generalization ability and scalability of DCIR.}

\subsection{Qualitive Results}

In Figure~\ref{fig:vis_example}, we visualize an episode of our agent performing the \textit{Cooperative Navigation~(5v0)} task. During the process, at time step $t = 5$, both \textit{Agent1}\textcolor[RGB]{112,173,71}{$\bullet$} and \textit{Agent2} \textcolor[RGB]{255,137,124}{$\bullet$} cover the same goal and \textit{Agent1} performs a ``humble'' behavior to move towards another one while \textit{Agent2} chooses to ``stay''. This eventually leads to two agents covering different goals without conflict~(time step $t = 10$). At time step $t = 14$, \textit{Agent3} \textcolor[RGB]{51,187,238}{$\bullet$} encounters the same situation. At this time, the intrinsic reward we proposed should encourage \textit{Agent3} and \textit{Agent4} \textcolor[RGB]{144,97,167}{$\bullet$} to behave inconsistently, which means $\alpha^3_4$ in the Equation~\ref{eq:dair} should be a positive value. \textit{Agent3} also should be rewarded for maintaining consistent behavior with \textit{Agent1} to perform ``humble'' behavior. To this end, $\alpha^3_1$ should be negative. We input the state of time step $t = 14$ into the DSN of \textit{Agent3} and find that the output for $\alpha^3_4$ is $0.28$ and  $\alpha^3_1$ is $-0.05$, suggesting the correctness of DSN. 

Ultimately, \textit{Agent3} adopts a behavior consistent with \textit{Agent1} to explore another goal, while being inconsistent with \textit{Agent4}'s behavior. We calculate its consistency at the time step $t = 14$ according to the Equation~\ref{eq:kl}, and get its consistency $\mathcal{C}$ with \textit{Agent1} and \textit{Agent4} to be $0.52$ and $5.57$, respectively. Notably, in our method, the smaller $\mathcal{C}$ is, the more consistent the behavior of the two agents is. This shows that behavior consistency by our definition is reasonable.

\section{Conclusion}
\label{CL}

In this paper, we address the MARL cooperation problem from the perspective of behavior consistency between agents. We propose to assign each agent a dynamic consistency intrinsic reward~(DCIR) to enhance performance in multi-agent tasks. 
Experimental results show that the proposed method significantly outperforms baseline MARL methods on the Multi-agent Particle, Google Research Football and StarCraft II benchmark environments. Qualitative results also show that the proposed metric correctly defines the behavior consistency and the DSN successfully outputs the appropriate scale factors for incentivizing the agents to perform their optimal behaviors. 
 DCIR is still limited to inferring the behavior in an implicit way.
 In scenarios with more high-level and complex actions, interpretable explicit behaviors may be needed to help the environment give more accurate intrinsic rewards.
Future work will focus on better representing and modeling teammate behavior for agents. 

\newpage
\nocite{langley00}

\bibliography{example_paper}
\bibliographystyle{icml2024}

\newpage
\appendix
\onecolumn

\icmltitle{\Large{\textbf{Supplementary Materials for \\ ``DCIR: Dynamic Consistency Intrinsic Reward for \\ Multi-Agent Reinforcement Learning''}}}

In the supplementary, we provide more implementation details and experimental results of our DCIR method. We organize the supplementary as follows.

\begin{itemize}
    \item In Section \textbf{``Optimizing Method of DSN''}, we provide the formulaic derivation of the optimizing method of DSN.
    \item In Section \textbf{``More Implementation Details''}, we provide more implementation details of our method, including the model architectures and hyperparameters.
    \item In Section \textbf{``Symmetry-Breaking Experiments''}, we provide the experimental results under \textit{Symmetry-Breaking} setting.
    \item In Section \textbf{``Occupancy/collision Results and Agent-to-target/adversary Distance Results''}, we provide more results \wrt occupancy/collision and agent-to-target/adversary distance.
    \item In Section \textbf{``Pilot Study''}, we provide the details of the pilot study mentioned in Section~\ref{Intro}.
\end{itemize}

\section{Optimizing Method of DSN}
\label{sec:Opt_DSN}
The dynamic scale network~(DSN) is parameterized by $\eta$. As aforementioned, each agent has its own DSN to output the dynamic scaling factor $\alpha$ for behavior consistency with other team agents. The parameters of DSN should be optimized in the direction of increasing extrinsic rewards. To achieve this goal, we adopt an extra-level actor-critic framework as in LIIR~\cite{LIIR}. 

To streamline our analysis, we represent the policy $\boldsymbol{\pi}_{\theta_i}$ with parameters  $\theta_i$ for agent $i$. We adopt the Policy Gradiant~\cite{policy} method as the objective function of the policy in an extra-level actor-critic framework: 

\begin{equation}
J^\text{ex} = \mathbb{E}_{o_t, \boldsymbol{u}_t \sim \mD}\left[\log \boldsymbol{\pi}_{\theta_i}\left(\boldsymbol{u}^i_t \mid o^i_t\right) A^\text{ex}(o_t, \boldsymbol{u}_t)\right],
\end{equation}
here, the advantage $A^\text{ex}$ is estimated as $A^\text{ex}(o_t, \boldsymbol{u}_t)=r_\text{ex}(o_t, \boldsymbol{u}_t)+V^\text{ex}\left(o_{t+1}\right)-V^\text{ex}(o_t)$ following~\cite{LIIR,TRPO,PPO}. We denoted 
 $V^\text{ex}$ as the extrinsic value which is estimated by the extrinsic critic and $o_{t+1}$ as the next successive observations of the agents. 

 Given the updated policy $\boldsymbol{\pi}_{\theta_i^{\prime}}$ with parameters $\theta_i^{\prime}$ of agent $i$ and its own DSN (~\ie~$\operatorname{DSN}_{\eta_i}$), we use the chain rule to build the connection between $\eta_i$ and $J^\text{ex}$ as follows~\cite{LIIR}:

 \begin{equation}
\nabla_{\eta_i} J^{\text{ex}}=\nabla_{\theta_i^{\prime}} J^\text{ex} \nabla_{\eta_i} \theta_i^{\prime}
\end{equation}

where the first term $\nabla_{\theta_i^{\prime}} J^\text{ex}$ is formulated as
\begin{equation}
\nabla_{\theta_i^{\prime}} \log \boldsymbol{\pi}_{\theta_i^{\prime}}\left(\boldsymbol{u}_t^i \mid o_t^i\right) A^\text{ex}(o_t, \boldsymbol{u}_t)
\end{equation}
here we reuse the samples generated by $\theta_i$ with importance sampling method~\cite{important}.

As mentioned in Section~3.2, the updated parameter $\theta_i^{\prime}$ comes from the following stochastic gradient method:
\begin{equation}
\begin{aligned}
& \theta_i-\xi  \mathbb{E}_{o_t^i \sim \mD^i}\left[\nabla_{\theta_i}\mathbb{E}_{\boldsymbol{u}^i_t \sim \boldsymbol{\pi}_{\theta_i}\left(\left.\cdot\right|_t\right)}\left[\omega \log \boldsymbol{\pi}_{\theta_i}\left(\boldsymbol{u}^i_t \mid o_t^i\right)-\min _{\theta_i} Q_{\phi_i}^{Soft}\left(o_t^i, \boldsymbol{u}^i_t\right)\right]\right] \\
& \approx  \theta_i-\xi \mathbb{E}_{o_t^i \sim \mD^i} \left[\nabla_{\theta_i}\boldsymbol{\pi}_{\theta_i}\left(o_t^i\right)^\top\left[\omega \log \boldsymbol{\pi}_{\theta_i}\left(o_t^i\right)-Q_{\phi_i^{\prime}}^{Soft }\left(o_t^i\right)\right]\right]
\end{aligned}
\end{equation}
where $\xi$ is the learning rate, $\omega$ is the entropy temperature coefficient.

To this end, the second term $\nabla_{\eta_i} \theta_i^{\prime}$ is computed as

\begin{equation}
\begin{aligned}
& \nabla_{\eta_i} \theta_i^{\prime} \\
& =\mathbb{E}_{o_t^i \sim \mD^i}\nabla_{\eta_i}\left[\theta_i-\xi \nabla_{\theta_i}\left[\omega \boldsymbol{\pi}_{\theta_i}\left(o_t^i\right)^\top \log \boldsymbol{\pi}_{\theta_i}\left(o_t^i\right)-\boldsymbol{\pi}_{\theta_i}\left(o_t^i\right)^\top Q_{\phi_i^{\prime}}^{Soft }\left(o_t^i\right)\right]\right] \\
& =\mathbb{E}_{o_t^i \sim \mD^i}\nabla_{\eta_i}\left[-\xi \nabla_{\theta_i} \omega \boldsymbol{\pi}_{\theta_i}\left(o_t^i\right)^\top \log \boldsymbol{\pi}_{\theta_i}\left(o_t^i\right)+\xi \nabla_{\theta_i} \boldsymbol{\pi}_{\theta_i}\left(o_t^i\right)^\top Q_{\phi_i^{\prime}}^{Soft }\left(o_t^i\right)\right] \\
& =\mathbb{E}_{o_t^i \sim \mD^i}\nabla_{\eta_i}\left[\xi \nabla_{\theta_i} \boldsymbol{\pi}_{\theta_i}\left(o_t^i\right)^\top Q_{\phi_i^{\prime}}^{Soft }\left(o_t^i\right)\right] \\
& =\xi \mathbb{E}_{o_t^i \sim \mD^i} \nabla_{\theta_i} \boldsymbol{\pi}_{\theta_i}\left(o_t^i\right)^\top  \nabla_{\eta_i} Q_{\phi_i^{\prime}}^{Soft }\left(o_t^i\right) \\
& =\xi \mathbb{E}_{o_t^i \sim \mD^i} \nabla_{\theta_i} \boldsymbol{\pi}_{\theta_i}\left(o_t^i\right)^\top  \nabla_{\phi_i^{\prime}} Q_{\phi_i^{\prime}}^{Soft }\left(o_t^i\right)  \nabla_{\eta_i} \phi_i^{\prime}
\end{aligned}
\end{equation}

The parameter of proxy critic $i$ is derived as
\begin{equation}
\begin{aligned}
& \nabla_{\eta_i} \phi_i^{\prime} \\
& = \mathbb{E}_{o_t^i \sim \mD^i} \nabla_{\eta_i}\left[\phi_i-\xi \nabla_{\phi_i}\left(Q_{\phi_i}^{Soft }\left(o_t^i\right)-Q_{\text {target }}\right)^2\right] \\
& = - \mathbb{E}_{o_t^i \sim \mD^i} \nabla_{\eta_i}\left[2\left(Q_{\phi_i}^{Soft }\left(o_t^i\right)-Q_{\text {target }}\right) \xi \nabla_{\phi_i} Q_{\phi_i}^{Soft }\left(o_t^i\right) \right] \\
& = -2 \xi \mathbb{E}_{o_t^i \sim \mD^i} \nabla_{\eta_i}\left(Q_{\phi_i}^{Soft }\left(o_t^i\right)-Q_{\text {target }}\right)  \nabla_{\phi_i} Q_{\phi_i}^{Soft }\left(o_t^i\right) \\
& = 2 \xi \mathbb{E}_{o_t^i \sim \mD^i} \nabla_{\eta_i}Q_{\text {target }}  \nabla_{\phi_i} Q_{\phi_i}^{Soft }\left(o_t^i\right)
\end{aligned}
\end{equation}
where $Q_{\text {target }} = r_\text{ex}^{t}+\beta \times r_{\text{DCIR}}^{i, t} + \underset{u_{t+1}^i \sim \boldsymbol{\pi}_{\theta_i^{\prime}}}{\mathbb{E}} \left[Q_{\phi_i^{\prime}}^{Soft }\left(o_{t+1}^i\right)-\omega \log \left(\boldsymbol{\pi}_{\theta_i^{\prime}}\left(u_{t+1}^i \mid o_{t+1}^i\right)\right)\right]$


Therefore,
\begin{equation}
\nabla_{\eta_i} \phi_i^{\prime} = 2 \beta \xi \mathbb{E}_{o_t^i \sim \mD^i} \nabla_{\eta_i}r_{\text{DCIR}}^{i, t}  \nabla_{\phi_i} Q_{\phi_i}^{Soft }\left(o_t^i\right)
\end{equation}

Hence we have:
\begin{equation}
\begin{aligned}
& \nabla_{\eta_i} \theta_i^{\prime} \\
& =\xi \mathbb{E}_{o_t^i \sim \mD^i} \nabla_{\theta_i} \boldsymbol{\pi}_{\theta_i}\left(o_t^i\right)^\top  \nabla_{\phi_i^{\prime}} Q_{\phi_i^{\prime}}^{Soft }\left(o_t^i\right)  \nabla_{\eta} \phi_i^{\prime}\\
& = 2 \beta \xi^2 \mathbb{E}_{o_t^i \sim \mD^i} \nabla_{\theta_i} \boldsymbol{\pi}_{\theta_i}\left(o_t^i\right)^\top  \nabla_{\phi_i^{\prime}} Q_{\phi_i^{\prime}}^{Soft }\left(o_t^i\right)  \nabla_{\eta_i}r_{\text{DCIR}}^{i, t}  \nabla_{\phi_i} Q_{\phi_i}^{Soft }\left(o_t^i\right)
\end{aligned}
\end{equation}
where the $r_{\text {DCIR }}^{i, t}=\sum_{j \in \mathcal{N}(i)} \alpha^i_j \times \mathcal{C}_{i, j, t}$ and the $\alpha^i_j$ is parameterized by $\eta_i$. In this way, we can connect the parameter updating of DSN with the objective $J^\text{ex}$ to ensure the  dynamic scaling factor $\alpha$ to be optimized in the direction of increasing extrinsic rewards.

\section{More Implementation Details}
\label{sec:model_hyp}
We use multi-layer perceptron (MLP) to implement all the models, including our DSN models, actor models, and critic models. The replay buffer size is $1,000,000$ for both the multi-agent particle environment and the Google research football environment. The training batch size is $1024$ and $256$ for the multi-agent particle environment and google research football environment, respectively. We adopt the Adam optimizer for parameter learning. The learning rate of actor and critic is $0.001$ for the multi-agent particle environment while $0.0003$ for the Google research football environment. We set discount factors $0.95$ and $0.99$ in  the multi-agent particle and Google research football environments, respectively. The soft update coefficient is $0.01$ in the multi-agent particle environment and $0.005$ in the Google research football environment. The entropy temperature coefficient is fixed to be $0.1$ in the multi-agent particle environment while learnable in the Google research football environment initializing by $1.0$. As for the StarCraft II Micromanagement, the learning rate is $0.0005$ and the training batch size is $32$, and the other implementation settings keep the same as in LIIR~\cite{LIIR}. The above settings are the same for all baselines and our method across all tasks, as these are common parts. The hyperparameters of DSN differ in tasks, and we will detail them in the open-source code.

\section{Symmetry-Breaking Experiments}
\label{sec:SB}
In this section, we study the challenges of efficiently allocating the sub-goals in multi-agent collaboration~\cite{symmetry}. We consider the symmetry-breaking setting as in ELIGN~\cite{ELIGN}. Specifically, for \textit{Coop Nav.} and \textit{Hetero Nav.}, we initialize all the agents in the same position and draw the goals randomly around the agents on a circle perimeter with a certain radius, which is equal to the value of the world radius minus the greatest goal size. As for \textit{Phy Decep.}, we place both agents and adversaries at the origin and the landmarks randomly around the agents on a circle perimeter with a certain radius. In \textit{Pred-prey.} task, we initialize the agents in the same position while the adversaries are put randomly in a circle. The agents and adversaries in \textit{Keep-away.} are placed in the same way as in \textit{Pred-prey.}, and the landmarks are initialized on a circle perimeter. Symmetry-breaking setting raises a challenge for the efficient goal allocation of the agents without supervision.

In Table~\ref{tab:breaking-reward}, we show the test results under \textit{Symmetry-Breaking} setting in the multi-agent particle environment. The results show that DCIR significantly beats the baselines, suggesting that DCIR helps the agents to break the deadlock of the same initial position. Through behavior consistency coordination among the agents, each agent can find its own optimal sub-goal, thus completing the task efficiently. We also notice that DCIR does not outperform all the baselines in \textit{Phy Decep.}, we hypothesize that this is because this task starts with both the agent and the adversaries in the same position, leading to a phase where their observations align. During this, the agents not only need to perform the dynamic behavior consistency with teammates to occupy the goals but also to deceive adversaries without confusing teammates. This enormously increases the difficulty of behavior consistency decision-making.

\begin{table*}[!t]
\resizebox{1.0\linewidth}{!}{\begin{tabular}{lcccccccccccc}
\topline
\multirow{2}{*}{Methods} &  & \multicolumn{3}{c}{MPE~(Cooperative)}     &  & \multicolumn{5}{c}{MPE~(Competitive)}    \\   \cmidrule{3-5} \cmidrule{7-11} 
                         &  & Coop Nav. (5v0)      &  & Hetero Nav. (6v0)      &  & Phy Decep. (4v2)      &  &  Pred-prey. (4v4)        &  &  Keep-away. (4v4)                \\ \midline
SPARSE               &  &   $328.24\pm24.17$       &  &  $405.08\pm21.53$       &  &  $172.87\pm32.43$     &  & $-35.40\pm8.63$       &   & $1.37\pm3.48$ \\
EXP-self             &  &   $295.48\pm20.54$       &  &  $436.17\pm26.30$       &  &  $202.39\pm26.06$     &  & $-11.19\pm3.65$       &   & $9.24\pm8.49$ \\
EXP-team             &  &   $316.33\pm14.44$       &  &  $422.71\pm13.24$       &  &  $229.50\pm28.29$     &  & $-11.56\pm6.37$       &   & $-1.29\pm1.58$\\
ELIGN                &  &   $357.40\pm19.52$       &  & $417.94\pm22.29$        &  &  $184.21\pm23.16$     &  & $-7.34\pm5.12$        &   &  $18.71\pm14.78$ \\
LIIR                 &  &   $337.14\pm{10.01}$            &  &  $439.07\pm{25.17}$            &  &  $184.25\pm{58.98}$          &  & $-52.98\pm{13.55}$           &  &  $2.64\pm{17.86}$ \\
\midline
\textbf{DCIR(Ours)}  &  & $\textbf{360.46}\pm\textbf{9.65}$  &  & $\textbf{461.90}\pm\textbf{40.26}$    &  & $214.61\pm31.05$    &  &  $-\textbf{7.05}\pm\textbf{3.70}$        &  &  $\textbf{23.56}\pm\textbf{13.81}$ \\ \bottomline

\end{tabular}
}
\caption{Mean test episode extrinsic rewards and standard errors in multi-agent particle environment under \textit{Symmetry-Breaking} setting. Higher values are better.}
\label{tab:breaking-reward}

\end{table*}

\begin{table*}[!t]
\resizebox{1.0\linewidth}{!}{\begin{tabular}{lcccccccccccc}
\topline
\multirow{2}{*}{Methods} &  & \multicolumn{3}{c}{MPE~(Cooperative)}     &  & \multicolumn{5}{c}{MPE~(Competitive)}    \\   \cmidrule{3-5} \cmidrule{7-11} 
                         &  & Coop Nav. (5v0)~↑      &  & Hetero Nav. (6v0)~↑      &  & Phy Decep. (4v2)~↑      &  &  Pred-prey. (4v4)~↓        &  &  Keep-away. (4v4)~↑                \\ \midline
SPARSE               &  &   $0.37\pm0.05$ && $0.38\pm0.03$ && $0.75\pm0.04$ && $0.13\pm0.03$ && $0.04\pm0.02$\\
EXP-self             &  &   $0.29\pm0.03$ && $0.43\pm0.02$ && $0.66\pm0.06$ && $0.05\pm0.02$ && $0.12\pm0.08$\\
EXP-team             &  &   $0.32\pm0.02$ && $0.34\pm0.01$ && $0.78\pm0.07$ && $0.05\pm0.02$ && $0.02\pm0.01$\\
ELIGN                &  &   $0.37\pm0.03$ && $0.40\pm0.03$ && $0.96\pm0.20$ && $0.04\pm0.03$ && $0.06\pm0.03$ \\
LIIR                 &  &   $0.38\pm{0.01}$ &  &  $0.36\pm{0.03}$ &  &  $1.01\pm{0.25}$ &  & $0.22\pm{0.06}$ &  &  $0.46\pm{0.24}$ \\
\midline
\textbf{DCIR(Ours)}  &  & $0.38\pm0.02$  &  & $0.34\pm0.02$    &  & $0.57\pm0.05$    &  &  $0.03\pm0.01$        &  &  $0.15\pm0.08$ \\ \bottomline

\end{tabular}
}
\caption{Mean test episode occupancy~(for tasks except \textit{Pred-prey.}) and collision~(only for \textit{Pred-prey.}) per step and standard errors in multi-agent particle environment under \textit{Symmetry-Breaking} setting. }
\label{tab:breaking-goal}

\end{table*}

\begin{table*}[!t]
\resizebox{1.0\linewidth}{!}{\begin{tabular}{lcccccccccccc}
\topline
\multirow{2}{*}{Methods} &  & \multicolumn{3}{c}{MPE~(Cooperative)}     &  & \multicolumn{5}{c}{MPE~(Competitive)}    \\   \cmidrule{3-5} \cmidrule{7-11} 
                         &  & Coop Nav. (5v0)~↓     &  & Hetero Nav. (6v0)~↓      &  & Phy Decep. (4v2)~↓     &  &  Pred-prey. (4v4)~↑        &  &  Keep-away. (4v4)~↓               \\ \midline
SPARSE               &  &   $0.36\pm0.01$ && $0.42\pm0.02$ && $0.35\pm0.01$ && $2.04\pm0.18$ && $3.10\pm0.29$\\
EXP-self             &  &   $0.50\pm0.04$ && $0.37\pm0.02$ && $0.38\pm0.03$ && $2.35\pm0.12$ && $2.70\pm0.35$\\
EXP-team             &  &   $0.43\pm0.03$ && $0.45\pm0.01$ && $0.35\pm0.02$ && $2.39\pm0.15$ && $3.37\pm0.19$\\
ELIGN                &  &   $0.42\pm0.03$ && $0.41\pm0.02$ && $0.37\pm0.01$ && $2.25\pm0.15$ && $2.62\pm0.30$\\
LIIR                 &  &   $0.40\pm{0.04}$ &  &  $0.44\pm{0.03}$            &  &  $0.31\pm{0.02}$          &  & $3.16\pm{0.23}$           &  &  $4.08\pm{0.37}$ \\
\midline
\textbf{DCIR(Ours)}  &  & $0.42\pm0.02$  &  & $0.46\pm0.02$    &  & $0.41\pm0.04$    &  &  $2.52\pm0.12$        &  &  $2.98\pm0.32$ \\ \bottomline

\end{tabular}
}
\caption{Mean test episode agent-to-target (for tasks except for \textit{Pred-prey.}) distance and agent-to-adversary~(only for \textit{Pred-prey.}) distance per step and standard errors in multi-agent particle environment under \textit{Symmetry-Breaking} setting. }
\label{tab:breaking-dis}

\end{table*}

\section{Occupancy/collision Results and Agent-to-target/adversary Distance Results}
\label{sec:GO}

In this section, we provide more results on occupancy/collision and agent-to-target/agent-to-adversary distance, as illustrated in Table~\ref{tab:breaking-goal}, ~\ref{tab:breaking-dis}, ~\ref{tab:goal}, ~\ref{tab:dis}. Note that these metrics are for individual agents and are not necessarily directly related to team performance. For example, the occupancy of each agent is high while the overall reward is low, indicating that many agents may occupy conflicting goals.

\begin{table*}[!t]
\resizebox{1.0\linewidth}{!}{\begin{tabular}{lcccccccccccc}
\topline
\multirow{2}{*}{Methods} &  & \multicolumn{3}{c}{MPE~(Cooperative)}     &  & \multicolumn{5}{c}{MPE~(Competitive)}    \\   \cmidrule{3-5} \cmidrule{7-11} 
                         &  & Coop Nav. (5v0)~↑      &  & Hetero Nav. (6v0)~↑     &  & Phy Decep. (4v2)~↑      &  &  Pred-prey. (4v4)~↓        &  &  Keep-away. (4v4)~↑                \\ \midline
SPARSE               &  &   $0.50\pm0.04$ && $0.46\pm0.08$ && $1.20\pm0.10$ && $0.11\pm0.02$ && $0.08\pm0.02$\\
EXP-self             &  &   $0.48\pm0.03$ && $0.63\pm0.01$ && $1.20\pm0.08$ && $0.07\pm0.02$ && $0.15\pm0.06$\\
EXP-team             &  &   $0.53\pm0.03$ && $0.60\pm0.02$ && $1.20\pm0.09$ && $0.05\pm0.02$ && $0.06\pm0.01$\\
ELIGN                &  &   $0.56\pm0.04$          &  & $0.67\pm0.00$           &  &  $1.30\pm0.23$        &  & $0.04\pm0.02$         &  &  $0.10\pm0.02$ \\
LIIR                 &  &   $0.54\pm{0.01}$            &  &  $0.66\pm{0.00}$            &  &  $1.46\pm{0.23}$          &  & $0.23\pm{0.06}$           &  &  $0.36\pm{0.18}$ \\
\midline
\textbf{DCIR(Ours)}  &  & $0.57\pm0.01$  &  & $0.63\pm0.03$    &  & $1.34\pm0.01$    &  &  $0.05\pm0.02$        &  &  $0.20\pm0.09$ \\ \bottomline

\end{tabular}
}
\caption{Mean test episode occupancy~(for tasks except \textit{Pred-prey.}) and collision~(only for \textit{Pred-prey.}) per step and standard errors in multi-agent particle environment. }
\label{tab:goal}

\end{table*}

\begin{table*}[!t]
\resizebox{1.0\linewidth}{!}{\begin{tabular}{lcccccccccccc}
\topline
\multirow{2}{*}{Methods} &  & \multicolumn{3}{c}{MPE~(Cooperative)}     &  & \multicolumn{5}{c}{MPE~(Competitive)}    \\   \cmidrule{3-5} \cmidrule{7-11} 
                         &  & Coop Nav. (5v0)~↓      &  & Hetero Nav. (6v0)~↓      &  & Phy Decep. (4v2)~↓      &  &  Pred-prey. (4v4)~↑        &  &  Keep-away. (4v4)~↓                \\ \midline
SPARSE               &  &   $0.22\pm0.01$ && $0.27\pm0.05$ && $0.23\pm0.02$ && $2.03\pm0.15$ && $2.97\pm0.17$\\
EXP-self             &  &   $0.30\pm0.02$ && $0.21\pm0.01$ && $0.24\pm0.01$ && $2.18\pm0.13$ && $2.70\pm0.25$\\
EXP-team             &  &   $0.23\pm0.02$ && $0.22\pm0.01$ && $0.23\pm0.02$ && $2.29\pm0.12$ && $3.14\pm0.08$\\
ELIGN                &  &   $0.23\pm0.04$ && $0.19\pm0.00$ && $0.22\pm0.01$ && $2.12\pm0.16$ && $2.66\pm0.23$\\
LIIR                 &  &   $0.24\pm{0.02}$            &  &  $0.19\pm{0.00}$            &  &  $0.20\pm{0.01}$          &  & $3.14\pm{0.20}$           &  &  $4.23\pm{0.30}$ \\
\midline
\textbf{DCIR(Ours)}  &  & $0.24\pm0.03$  &  & $0.21\pm0.01$    &  & $0.21\pm0.00$    &  &  $2.40\pm0.16$        &  &  $2.89\pm0.17$ \\ \bottomline

\end{tabular}
}
\caption{Mean test episode agent-to-target/agent-to-target~(for tasks except \textit{Pred-prey.}) distance and agent-to-adversary~(only for \textit{Pred-prey.}) distance per step and standard errors in multi-agent particle environment. }
\label{tab:dis}

\end{table*}

\section{Pilot Study}
\label{sec:ps}
To inspect how severe the previous methods suffer in dynamic behavior consistency problem and how much DCIR alleviates this problem, we conduct two pilot studies.
We place 5 agents at the origin in MPE.
In \textbf{\textit{Study 1}}, we position one target at the origin, while the remaining 4 are randomly distributed around a circle centered at the origin, all within the observable range of agents. The agents need to behave \textbf{inconsistently} to go in different directions toward the targets.
In \textbf{\textit{Study 2}}, we place all the targets on the right of the origin within the observation range of the agents. In this case, agents need to maintain \textbf{consistent} behaviors to go right toward the target.
For Study 1, we report the proportion of agents that left the origin (with a denominator of 4 since one agent needs to stay at the origin which also has a target).
In Study 2, we report the proportion of agents that approach the targets. Each variant is repeated 1000 times in different seeds.

\vspace{-3mm}

\begin{table}[!ht]
\centering
\resizebox{0.4\linewidth}{!}{\begin{tabular}{lcccc}
\topline
        
                             &  & Study 1  &  & Study 2                   
\\ \midrule
ELIGN                &  &  $0.75\pm{0.07}$    &  &  $0.69\pm{0.18}$ \\  
DCIR(Ours)           &  &  $\textbf{0.88}\pm\textbf{0.08}$  &  &  $\textbf{0.97}\pm\textbf{0.02}$ \\ \midrule
\end{tabular}
}
\vspace{-3mm}
\caption{Pilot studies on dynamic behavior consistency}
\label{tab:con2}
\vspace{-4mm}
\end{table}

In Table~\ref{tab:con2}, compared with the existing SoTA method ELIGN, our delicately designed DCIR works well to encourage agents to perform dynamic behavior consistency at the right time and thus improve task performance.

\end{document}

%% file: def.tex
\def\mD{{\mathcal D}}

\DeclareMathAlphabet\mathbfcal{OMS}{cmsy}{b}{n}

\def\0{{\bf 0}}
\def\1{{\bf 1}}

\def\eg{\emph{e.g.}} 
\def\ie{\emph{i.e.}} 
 
 \def\vs{\emph{vs.}}
\def\wrt{{w.r.t.~}}

\usepackage{amsmath}